\documentclass[review]{elsarticle}

\usepackage{lineno,hyperref}
\modulolinenumbers[5]

\usepackage{amsmath}
\usepackage{times}
\usepackage{graphicx}
\usepackage{color}
\usepackage{multirow}
\usepackage{url}
\usepackage{comment}

\journal{Journal of NeuroComputing}

\graphicspath{{E:/Papers/elsarticle-template./Pictures/}}

\begin{document}

\begin{frontmatter}

\title{Analysis of Massive Heterogeneous Temporal-Spatial Data with 
	3D Self-Organizing Map and Time Vector }

\author{Yu Ding}
\address{Huazhong University of Science and Technology\\
	Luoyu Road 1037, Wuhan, China\\
	dingy@hust.edu.cn}
%\fntext[myfootnote]{tel +86 13006127295; fax +86 027-87540724; mail Luoyu Road 1037, Wuhan, China, post code 430074}


\begin{abstract}
Self-organizing map(SOM) have been widely applied in clustering, this paper focused on centroids of clusters and what they reveal. When the input vectors consists of time, latitude and longitude, the map can be strongly linked to physical world, providing valuable information. Beyond basic clustering, a novel approach to address the temporal element is developed, enabling 3D SOM to track behaviors in multiple periods concurrently. Combined with adaptations targeting to process heterogeneous data relating to distribution in time and space, the paper offers a fresh scope for business and services based on temporal-spatial pattern.
\end{abstract}


\end{frontmatter}

\linenumbers

\section{Introduction}
\subsection{Background of Research}
With the development of information gathering technology, people can access to tremendous amount of real-time occurrence data consisting of coordinates both in time and space, such as the e-commerce orders, Uber requests\cite{Uberrequest}, crime incident reports\cite{7MajorFelony}, and vehicle collisions\cite{carcollision}. The massiveness conceals patterns requiring feasible a tool to identify. Following research is tightly related to their features listed below: 
\begin{itemize}
	\item[1)] Extremely dense. \\
	Density in both time and space makes it impossible to track every input, therefore, determining centroids that represent the cluster they belong is essential for providing service based on time point and location, such as optimized warehouse site for delivery, effective patrol schedule and so forth.
	\item[2)] Multi-Periodically Structured. \\ 
	Data repeats in days, weeks and months, demanding us to analysis under multiple period. For example, vehicle collision's distribution fluctuates within 24 hours and days in one week.
	\item[3)] Heterogeneous.\\ 
	Inputs include both numeric and categorical data. The crime incidents data not only covers time and geographic variables, but also classification of felonies. How to incorporate heterogeneous data is challenging yet useful. 
\end{itemize}

The research was conducted in a progressive manner, beginning with basic clustering on time, latitude and longitude, aiming to solve the problem 1. Since not much prior experiment has been done to this type of data, making it necessary to test with different configurations and parameters to attain better accuracy. Next, the highlight of this passage, time vector is introduced to handle problem 2. Finally, adaptations on SOM itself are made to overcome the difficulties mentioned in 3.  

\subsection{Introduction on algorithm}
Self-organizing map\cite{58325} is an unsupervised and efficient algorithm for clustering, which not only allows people to divide the data into sectors, but also to understand their topographic relation. Following terminology is introduced with potential application. 
\begin{itemize}
	\item Nodes, building up self-organizing map's grid, the centroids of clusters.
	\item Neighborhood, a sector within certain radius centering around a selected nodes.
	\item Codebook Vector, indicating where a node is situated in coordinate's of input data. Its value offers detailed information about the nodes, like when and where is most representative for a cluster of crimes.
	\item Best Matching Unit(BMU), the node with least distance to a chosen input. BMU can be regarded as location with least cost. For instance, a driver at BMU has the least distance to a customer.   
	\item Hits, inputs that belongs to a specific cluster, like the recipients that a warehouse served. 
\end{itemize}

\section{Clustering with 3D Self-Organizing Map}\label{Basic}
\subsection{Construction of SOM}
The framework of 3D grid consists of layers covering latitude and longitude axises, while different layers array along time axis. Nodes in one layer are marked with same color.
\begin{figure}[!h]
	\begin{center}
		\label{3dgrid}
		\includegraphics[width=4.8in]{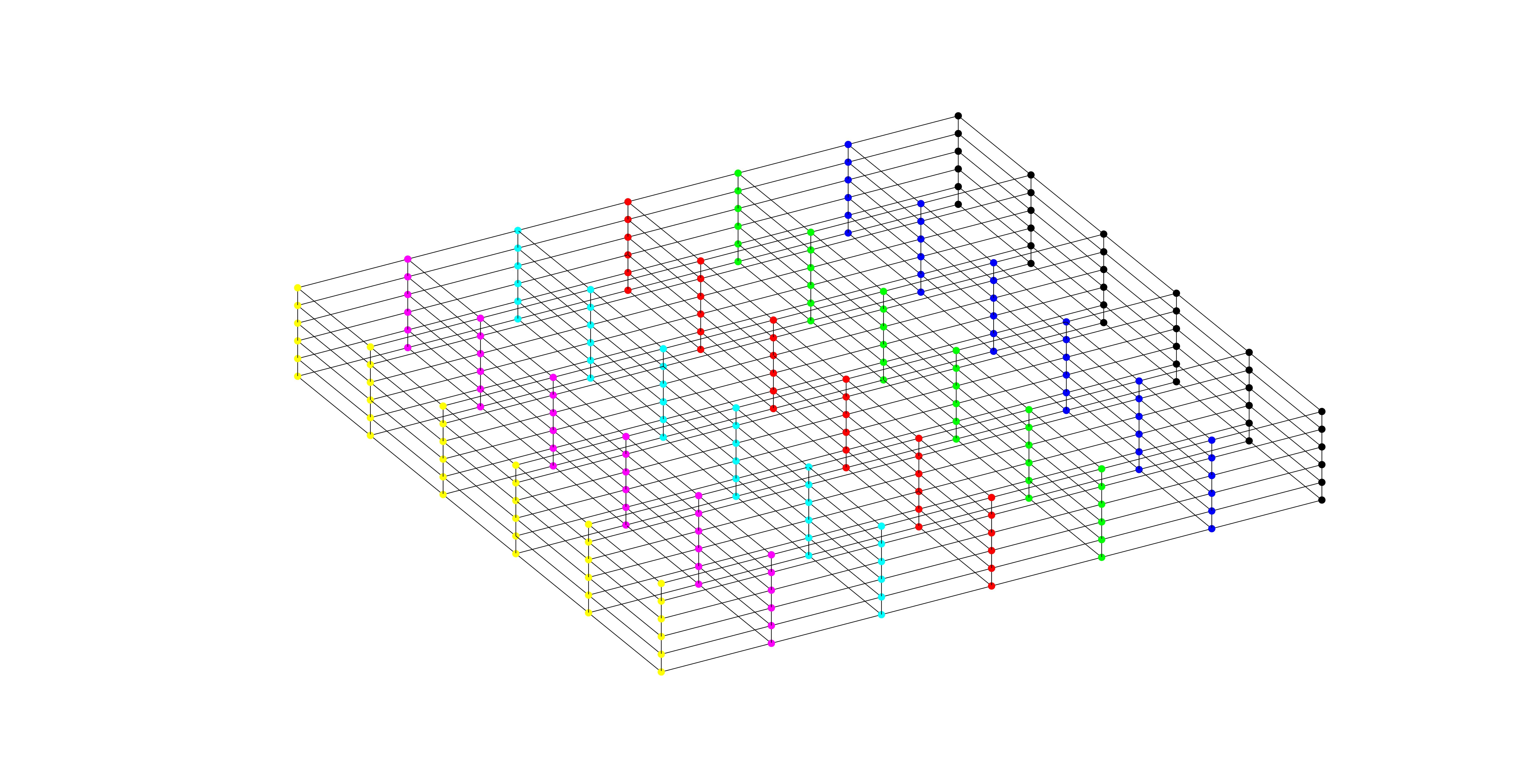}
		\caption{ A $7\times6\times6$ 3D SOM Grid}
	\end{center}
\end{figure}
\subsubsection{Significance of Pre-Processing Input}
According to previous research, normalization will promote the quality of map\cite{demartines1992kohonen}. This paper compared two types of normalization, rescaling to [0,1] and standardization with z-score, both of which improves map's quality by drastically reducing errors introduced later. Detailed results is available in \nameref{A1}.
\subsubsection{Initialization}
In the research it's found that in order to achieve a well-organized map, grid must be initialized linearly and the side that parallels to time axis should be initialized with greatest number of nodes, in other words, the number of layers should be larger than each side of layers. Moreover, initializing the map with different sizes yields an interesting yet telling phenomenon that if lengths of grid's sides are not selected properly, the map will rotate during reiterations, ending up with poor quality. Explanation is discussed in \nameref{A2}.

\subsubsection{Training Progress}
After initializing, training of SOM is summarized as following: 
\begin{itemize}
	\item[1)] Find BMU for every input under Euclidean metric;
	\item[2)] Update the map in batch mode;
	\item[3)] Reiterate from step one;
\end{itemize}
Since this research engages with heavy calculation, batch training is more preferable than sequential one, and using voronoi set can accelerate updating progress further\cite{batch} with following function.
\begin{equation}
	m_{i} =\dfrac{\sum\limits_{j} n_{j}h_{ci}x(t)}{\sum\limits_{j} n_{j}h_{ci}}
\end{equation}
On the left, $m_{i}$ is a codebook vector, and $n_{j}$ is number of hits for node j, $h_{c,i}$ measures influence or weigh of node c to node i, determined by Gaussian neighborhood function $ h_{c,i} = e ^ {-\frac{dm(m_{c},m_{i})^2} {r^2}} \label{gaussian}$. 

\subsection{Measurement of Quality}
\subsubsection{Quantization Error}
Quantization error(QE) measures average distance between inputs to their BMUs. 
\begin{equation}
	QE = \dfrac{\sum\limits_{i=1}^{n} d(x_{i},m_{j})}{n}
\end{equation}
$d(x_{i},m_{j})$ is the distance of between input $x_{i}$ and its best matching unit $m_{j}$ and $n$ is the total number of input. QE is important in application, for instance, selecting warehouse according to map with small QE can be economical for shipping industry.

\subsubsection{Topographic Error}
Topographic error(TE) measures the portion of input whose BMU and second BMU connected directly in grid.
\begin{equation}
	TE = \dfrac{\sum\limits_{i=1}^{n} \delta(x_{i})}{n}
\end{equation}
The value of $\delta(x_{i})$ is determined by whether the nodes has second minimum distance to input$x_{i}$ is a neighbor to BMU. If so, $\delta(x_{i})=0$, otherwise it will be set as 1. If a map exhibits low TE, while the BMU's site is not available in application like police station, second BMU might be good substitute. 

\subsection{Reliability Estimation}
\subsubsection{Generating Correlation Matrix}
In order to fully evaluate map's soundness in application, we need to ascertain whether centroids and their hits distribute correspondingly to density of input. It can be achieved by converting input's regional density into frequencies as following steps:
\begin{itemize}
	\item[1)] Slicing the input space into cubics;
	\item[2)] Generating a matrix contains numbers of inputs in each cubic. Precision can be set via choosing different pieces of slices;
	\item[3)] Counting the centroids and hits in every cubic, obtaining other two matrices with same dimension;
	\item[4)] Measuring their correlations to input frequency matrix. 
\end{itemize}
If a map is representative, correlation coefficient(COR) should approximate to 1.
\begin{equation*}
	COR(X,Y) = \frac{\sum\limits_{j=1}^m\sum\limits_{i=1}^n (x_{ij}-\bar{X})(y_{ij}-\bar{Y})}
	{\sqrt{\sum\limits_{j=1}^m\sum\limits_{i=1}^n (x_{ij}-\bar{X})^2 \sum\limits_{j=1}^m\sum\limits_{i=1}^n (y_{ij}-\bar{Y})^2}}
\end{equation*}
where $\bar{X}$ and $\bar{Y}$ are mean of matrices $X$,$Y$. 

\subsubsection{Projection}
Projecting high-dimensional into subspaces makes interpretation of interior relations more accessible. 
In clustering input with time, latitude and longitude, under the notion of projection, estimation can be divided into two parts:
\begin{itemize}
	\item Spatial consistency, in fixed durations, measuring how distribution of nodes and their hits match with input. To be more concrete, it measures frequency matrix slices along the time axis.
	\item Temporal coherence, measuring if spatial consistency remain stable in different periods, by setting the durations to other periods.
\end{itemize}
In further analysis, projections is essential for assessing performance. While probating multiple period patterns, projecting data into different temporal dimensions allows us to observe variance under periods of varied durations. When input covers numeric and categorical variables, we need to evaluate accuracy of both temporal-spatial clustering and category classification. \\

\section{Exploring the Potential of 3D SOM}\label{Explore}
Apart from identify centroids in time-latitude-longitude space, basic model is insufficient for demands in real world. First, only distribution in 24 hours is considered, whereas weekly or monthly fluctuation is ignored. In addition, input data comprises numeric elements and categorical ones, like name of purchased item, felony classifications, among which distance cannot be reckoned via Euclidean norm. Further analysis will investigate two possible solution to overcome these limitations.

\subsection{Broadening Timeline with Time Vector}
If time is studies as one dimension line while every moment is a zero-dimensional point, details in shorter spans are compressed when studying long term patterns. To circumvent this paradox, time point can be converted into time vector:
\begin{equation}
	t(\frac{n_{i}}{P_{i}}\times\cdots\times\frac{n_{j}}{P_{j}}) \Rightarrow 
	t^{*}<\frac{n_{i}}{P_{i}},\cdots, \frac{n_{j}}{P_{j}}>
\end{equation}
On the left side, time $\displaystyle t$ is expressed as a product of fractions where the denominators $\displaystyle P_{i}$ are different periods, for day, week, and month are 24*60, 7 and 12 respectively. Numerators $\displaystyle n_{i} $ are sequences in period $\displaystyle P_{i}$. The right side is a vector consisting of multipliers on the left. For example, 8:30 on Tuesday in March would be $(\displaystyle\frac{510}{1440}\times \frac{2}{7}\times\frac{3}{12}) \rightarrow <\displaystyle\frac{510}{1440},\displaystyle\frac{2}{7}, \displaystyle\frac{3}{12}>$.

In this passage, 2D time vector is used, turning input into 4D vectors(day-week-latitude-longitude). Combined with 3D SOM, this technique empowers us to inspect multiple periods' behaviors simultaneously. Furthermore, it could reveal influence between temporal periods, for instance, as days elapse in a week, when peaks occur in one day will also change. 

\subsection{Mixed with Categorical Data}
The major difficulty in analysis of mixed data is to find a method determines the distance between variables, and update the map. In this paper, straightforward approach is employed to deal with heterogeneous data, introduced below:
\begin{itemize}
	\item[1)] Assigning the name strings with ID numbers arbitrarily, since we do not relay on their quantitative meaning, then input $x_{i}$ becomes a 4D vectors(time-latitude-longitude-ID); 
	\item[2)] Transforming $x_{i}$'s ID number $j$ into a binary vector $C_{x_{i}}(j)$, index of column with 1 indicate ID number it holds, $x_{i}$ now has two parts, 3D numeric vector$N_{x_{i}}$ and kD category vector $C_{x_{i}}$;
	\begin{equation}
		C_{x_{i}}(j) = <\overset{1}{0},\cdots\overset{j}{1}, \cdots\overset{k}{0}>_{k} 
	\end{equation}
	$k$ is number of categories.
	\item[3)] Total distance $D(x_{i},m_{j})$ between input $x_{i}$ and $m_{j}$ is sum of numeric variables $Dn(N_{x_{i}},N_{m_{j}})$ calculated under Euclidean norm and categorical part $Dc(C_{x_{i}},C_{m_{j}})$ deduced via logic operation \textit{AND}, if$x_{i}$, $m_{j}$share same ID number, set it to 0, if not, to 1;
	\begin{eqnarray}
		D(x_{i},m_{j})  & = & Dn(N_{x_{i}},N_{m_{i}}) + \alpha \times Dc (C_{x_{i}},C_{m_{i}})\\
		Dc(C_{x_{i}},C_{m_{j}})   & = & C_{x_{i}}\&C_{m_{j}}
	\end{eqnarray}
	$\alpha$ is a parameter to offset scale dominance of either $Dn(x_{ik},m_{jk})$ or $Dc(x_{ik},m_{jk})$ in next step; 
	\item[4)] Searching for BMUs, working out a $m \times k $ weight matrix \textbf{W}, $W_{ij}$ represent how category j weighs in nodes i. For details on \textbf{W}, please view \nameref{B1}.
	\item[5)] Applying winner-take-all strategy on \textbf{W}, codebook vector $m_{i}$'s ID variable is assigned with column $j$ that weighs most in row $W_{i}$.
	\begin{equation}
		m_{i} = \underset{j}{arg max}(W_{i}(j))
	\end{equation}
\end{itemize}
Before utilize it, we should be aware that binary coding process neglects inner structure of categorical variables, which is not suitable category classifications have affiliations\cite{hsu2012visualized}.

\section{Experimental Results}
Running the algorithms on two data sets, crime incidents and vehicle collisions, entering data in a single month at one time. In general, performances didn't vary much in input from different months, and crime incidents in \textbf{ Jan.2015} is selected for illustration, containing 7816 crime reports with date, time, latitude, longitude values and felony names. The CORs are affected by number of cubics dividing original space, for the impartiality of results, CORs are reckoned under a fixed division. Followings are result of $13 \times 8 \times 7$ map, measured under $8 \times 5 \times 5 $ cubics in time-latitude-longitude space. 
\subsection{Performance on Basic Clustering}
In different months, overall CORs between frequency matrices of nodes and input fluctuates around \textbf{0.82}, for CORs of hits and input hold steady above \textbf{0.95}. \\
Table \ref{tab:1} shows CORs during eight sections in 24 hours of sample mentioned above.
\begin{table}[h]
	\renewcommand{\arraystretch}{0.7}
	\caption{Coefficients between Input and Map}\label{tab:1}
	\begin{center}
		\begin{tabular}{|c|c|c|c|c|c|c|c|c|}
			\hline
			Input \& Nodes & 0.94 & 0.43 & 0.84 & 0.72 & 0.96 & 0.63 & 0.95 & 0.64 \\
			Input \& Hits  & 0.99 & 0.98 & 0.98 & 0.99 & 0.99 & 0.99 & 0.99 & 0.97 \\ \hline
		\end{tabular}
	\end{center}
\end{table}

While CORs offer a percentage-like assessment, table \ref{tab:2} provides a quantitative reflection of consistency between map and input. It sums up numbers in different sections.\\    
\begin{table}[!h]
	\renewcommand{\arraystretch}{0.7}
	\caption{Numbers in Each Section}\label{tab:2}
	\begin{center}
		\begin{tabular}{|c|c|c|c|c|c|c|c|c|c|}
			\hline
			Input & 976 & 540 & 729 & 1041 & 1239 & 1380 & 1244 & 667 & $R^{2}$ \\
			Nodes & 89  & 72  & 73  & 103  & 106  & 121  & 111  & 43  &  0.81 \\
			Nits  & 966 & 577 & 663 & 1121 & 1201 & 1387 & 1280 & 621 &  0.98\\ \hline
		\end{tabular}
	\end{center}
\end{table}\\
The last column contains the \textit{$R^{2}$} value deduced from linear regression to measure map's consistence to input. In order to observe in a perceptive manner on how map represent the input, heat map of input's density is painted with nodes in same section. 
\begin{figure}[!h]
	%\nopagenumber
	%\renewcommand{\baselinestretch}{1.0}
	\hfill
	\begin{center}
		\includegraphics[width=4.8in]{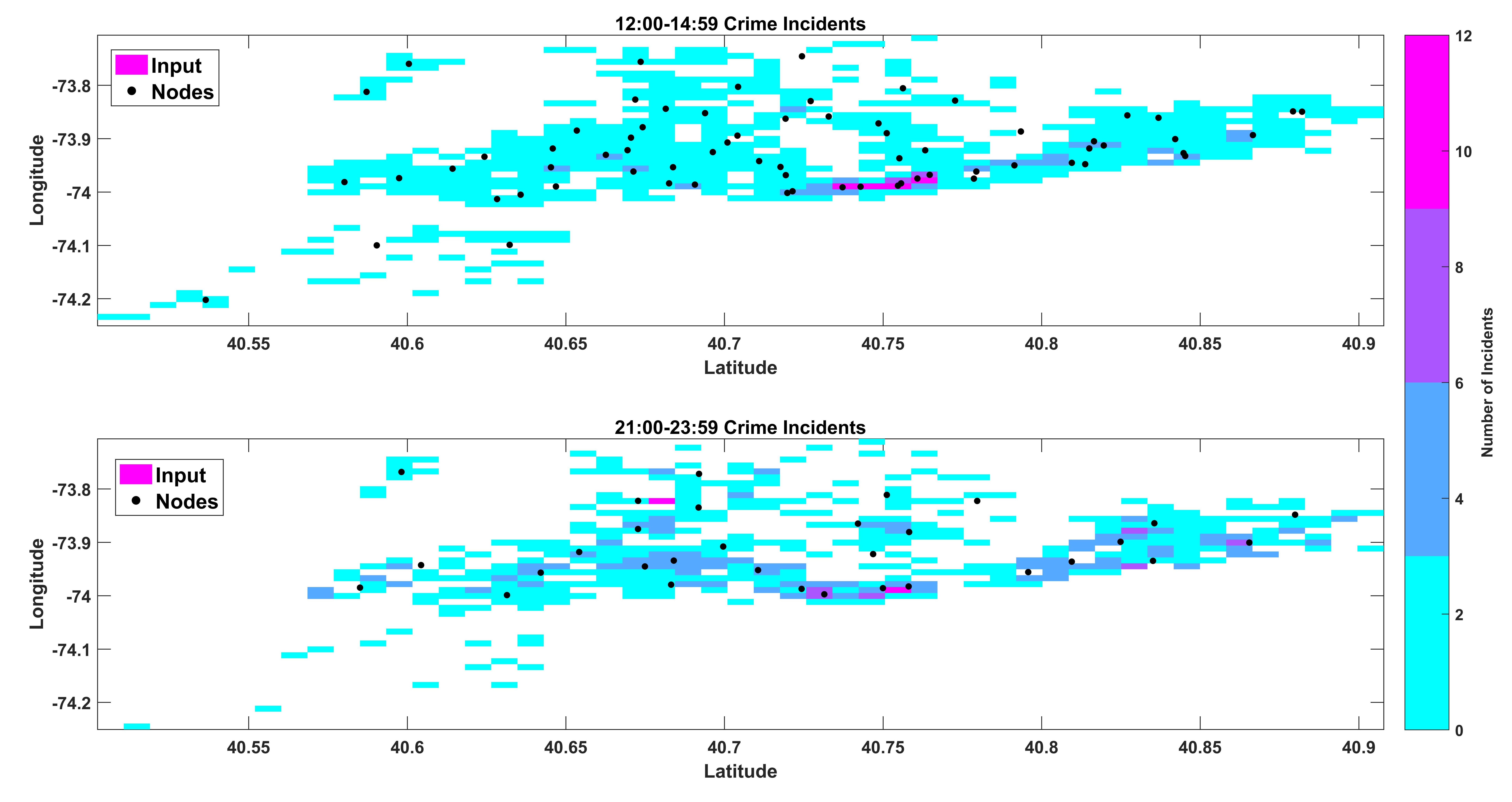}
		\caption{\label{p2} Daily Crime Incident Map in Jan.2015}
	\end{center}
	%\label{p2}
\end{figure}\\
In Figure \ref{p2}, two distinguishable intervals are selected to show how nodes match with the density of input. 

\subsection{Performance on Multi-Period Analysis}
2D time vector $<\frac{time}{24 \times 60},\frac{day}{7}>$ is used in studying daily and weekly behaviors of crime incidents. Assessing the reliability with projections into subspaces of day-latitude-longitude and week-latitude-longitude. Extra test was conducted on vehicle collisions, result is supplied in \nameref{C1}. \\
In study the daily pattern, table \ref{tab:3} represents CORs in eight sections of 24 hours.
\begin{table}[!h]
	\renewcommand{\arraystretch}{0.7}	
	\caption{Coefficients between Input and Map in Day}\label{tab:3}
	\begin{center}
		\begin{tabular}{|c|c|c|c|c|c|c|c|c|}
			\hline
			Input \& Nodes & 0.74 & 0.11 & 0.89 & 0.61 & 0.76 & 0.96 & 0.68 & 0.53   \\
			Inputs \& Hits & 0.97 & 0.97 & 0.95 & 0.98 & 0.99 & 0.96 & 0.97 & 0.98   \\ \hline
		\end{tabular}
	\end{center}
\end{table}

Sum up numbers in each section and apply linear regression to obtain $R^{2}$ value.
\begin{table}[!h]
	\renewcommand{\arraystretch}{0.7}	
	\caption{Daily Sum of Input and Map}
	\begin{center}
		\begin{tabular}{|c|c|c|c|c|c|c|c|c|c|}
			\hline
			Input & 976 & 540 & 729 & 1041 & 1239 & 1380 & 1244 & 667 & $R^{2}$ \\
			Nodes & 85  & 62  & 68  & 102  & 102  & 135  & 111  & 63  &  0.92   \\
			Hits  & 942 & 582 & 696 & 1100 & 1162 & 1456 & 1183 & 695 &  0.96   \\ \hline
		\end{tabular}
	\end{center}
\end{table}\\
In week-latitude-longitude subspace, CORs are measured day by day.
\begin{table}[!h]
	\renewcommand{\arraystretch}{0.7}	
	\caption{Coefficients between Input and Map in Week}
	\begin{center}
		\begin{tabular}{|c|c|c|c|c|c|c|c|}
			\hline
			Day in Week   & Mon. & Tue. & Wed. & Thur. & Fri. & Sat. & Sun. \\
			Input \& Nodes & 0.97 & 0.91 & 0.92 & 0.67  & 0.93 & 0.54 & 0.94 \\
			Inputs \& Hits & 0.99 & 0.99 & 0.99 & 0.98  & 0.99 & 0.99 & 0.99 \\ \hline
		\end{tabular}
	\end{center}
\end{table}

Then sum the numbers in each day of week, and calculate weekly \textbf{$R^{2}$} value.
\begin{table}[!h]
	\renewcommand{\arraystretch}{0.7}	
	\caption{Weekly Sum of Input and Map}
	\begin{center}
		\begin{tabular}{|c|c|c|c|c|c|c|c|c|}
			\hline
			Day in Week & Mon. & Tue. & Wed. & Thur. & Fri. & Sat. & Sun. &  \\
			Input    & 1412 & 1383 & 1184 &  921  & 1064 & 913  & 939  & $R^{2}$ \\
			Nodes    & 111  & 112  & 105  &  101  & 105  &  98  &  96  &  0.89   \\
			Hits     & 1417 & 1380 & 1181 &  918  & 1068 & 890  & 962  &  0.99   \\ \hline
		\end{tabular}
	\end{center}
\end{table}
\\
\\
\\
\\
\\
\\

Distribution on Tuesday and Thursday are selected to show how the map trace weekly variance.
\begin{figure}[!h]
	%\nopagenumber
	%\renewcommand{\baselinestretch}{1.0}
	\hfill
	\begin{center}
		\includegraphics[width=4.8in]{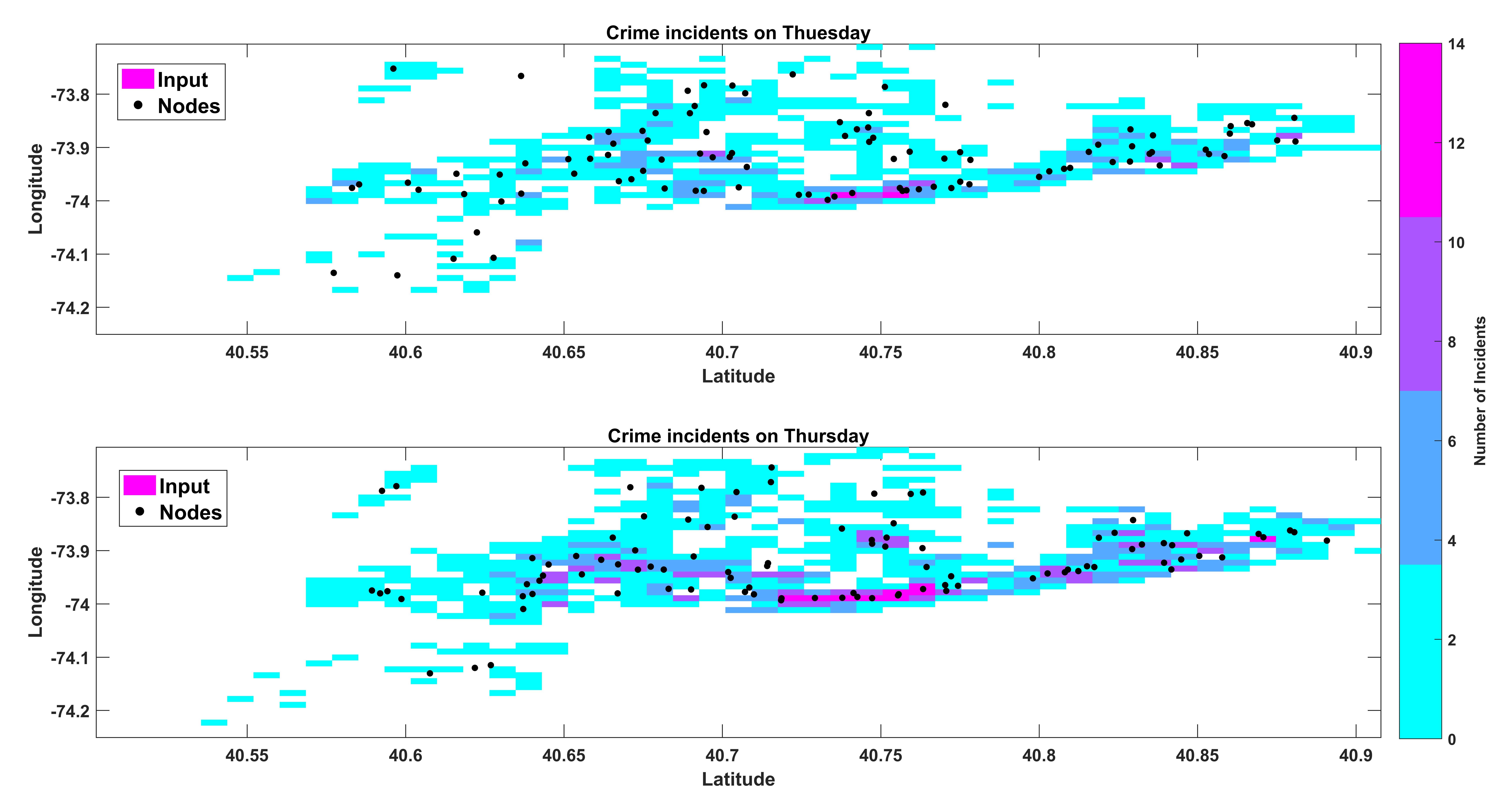}
		\caption{\label{fig3} Heat Map and Distribution of Nodes}
	\end{center}
	
\end{figure}

\begin{figure}[!h]
	%\nopagenumber
	%\renewcommand{\baselinestretch}{1.0}
	\hfill
	\begin{center}
		\includegraphics[width=4.8in]{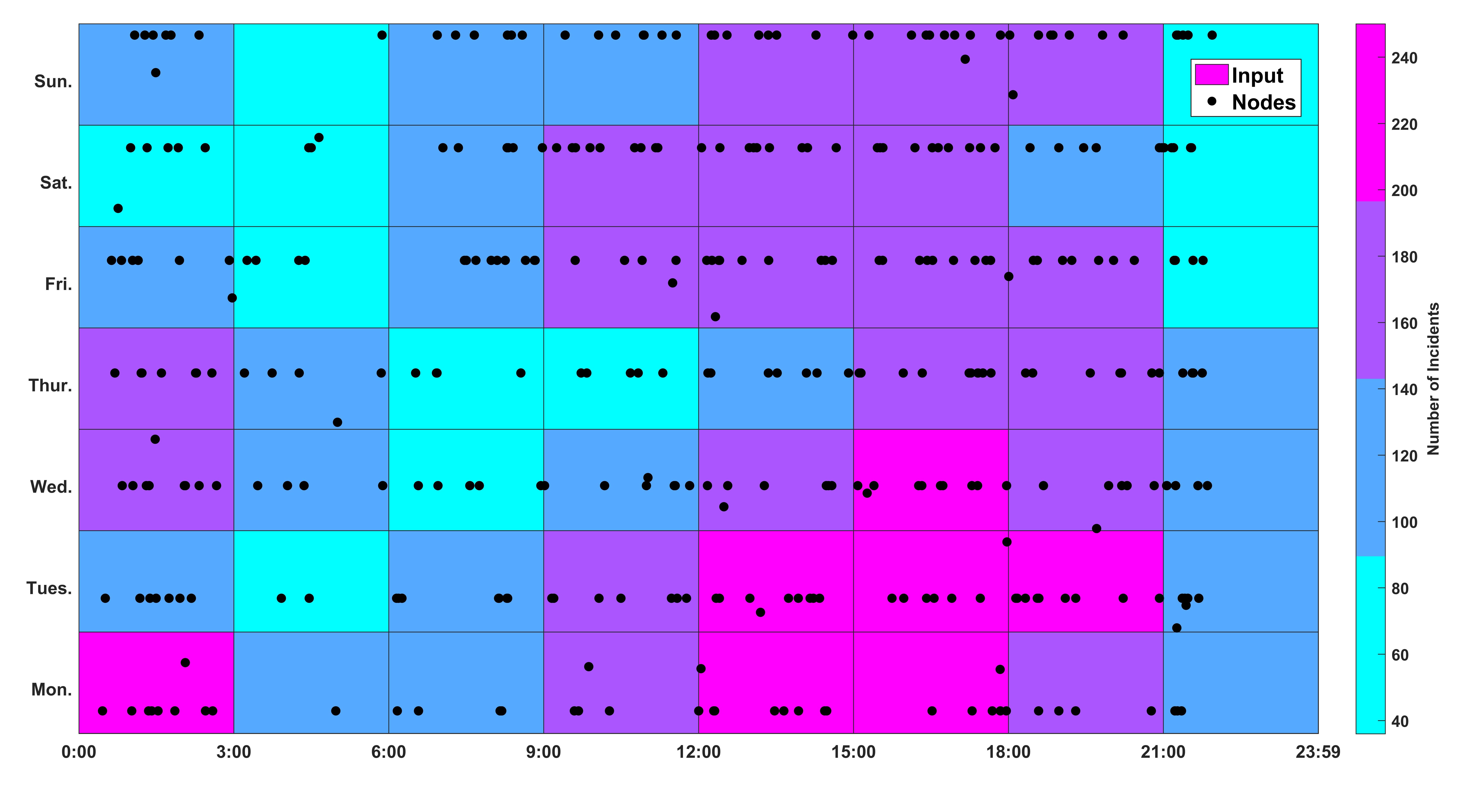}
		\caption{\label{Fig:weekday} Heat Map and Distribution of Nodes}
	\end{center}
	
\end{figure}

As mentioned, projections into day-week plane suggest temporal elements are interrelated, that behavior in 24 hours is affected by which day it is in week, where COR between input and map reaches \textbf{0.64}. Figure \ref{Fig:weekday} is heat map of incidents in day-week plane.\\
\\

\subsection{Performance on Heterogeneous Input}
The modified 3D SOM is tested with crime incidents data including felony types. Likewise, performance is gaged by calculate CORs under controlled felony IDs or fixed temporal intervals. To avoid monotony in representation, only results on clustering different IDs are provided here. Felony classification ID and temporal-spatial clustering of all input is supplied in \nameref{C2}.
\begin{table}[!h]
	\renewcommand{\arraystretch}{0.7}
	\caption{CORs of Different Felonies}
	\begin{center}
		\begin{tabular}{|c|c|c|c|c|c|c|c|}
			\hline
			ID Number &   1   &  2   &  3   &  4   &  5   &  6   &  7   \\ \hline
			Nodes   & -0.22 & 0.94 & 0.54 & 0.96 & 0.80 & 0.17 & None \\
			Hits    & 0.06  & 0.97 & 0.98 & 0.99 & 0.98 & 0.92 & None \\ \hline
		\end{tabular} 
	\end{center}
\end{table}
\\
\\
\\
\\

In same manner, sum up numbers in each type of crime and calculate \textbf{$R^{2}$}.
\begin{table}[!h]
	\renewcommand{\arraystretch}{0.7}
	\caption{Numbers in Each ID}
	\begin{center}
		\begin{tabular}{|c|c|c|c|c|c|c|c|c|}
			\hline
			ID Number  & 1  &  2   &  3   &  4   &  5   &  6  & 7  &  \quad    \\ \hline
			Input Data  & 94 & 1174 & 1374 & 3188 & 1438 & 510 & 38 & $R^{2}$   \\
			Layer Nodes & 5  & 125  & 139  & 312  & 104  & 43  & 0  &  0.98     \\
			SOM Hits   & 94 & 1182 & 1383 & 3204 & 1441 & 512 & 0  &  0.99     \\ \hline
		\end{tabular} 
	\end{center}
\end{table}

 Picture below shows heat maps of two selected types of crime and distribution of nodes with same ID.
\begin{figure}[!h]
	%\nopagenumber
	%\renewcommand{\baselinestretch}{1.0}
	\hfill
	\begin{center}		
		\includegraphics[width=4.8in]{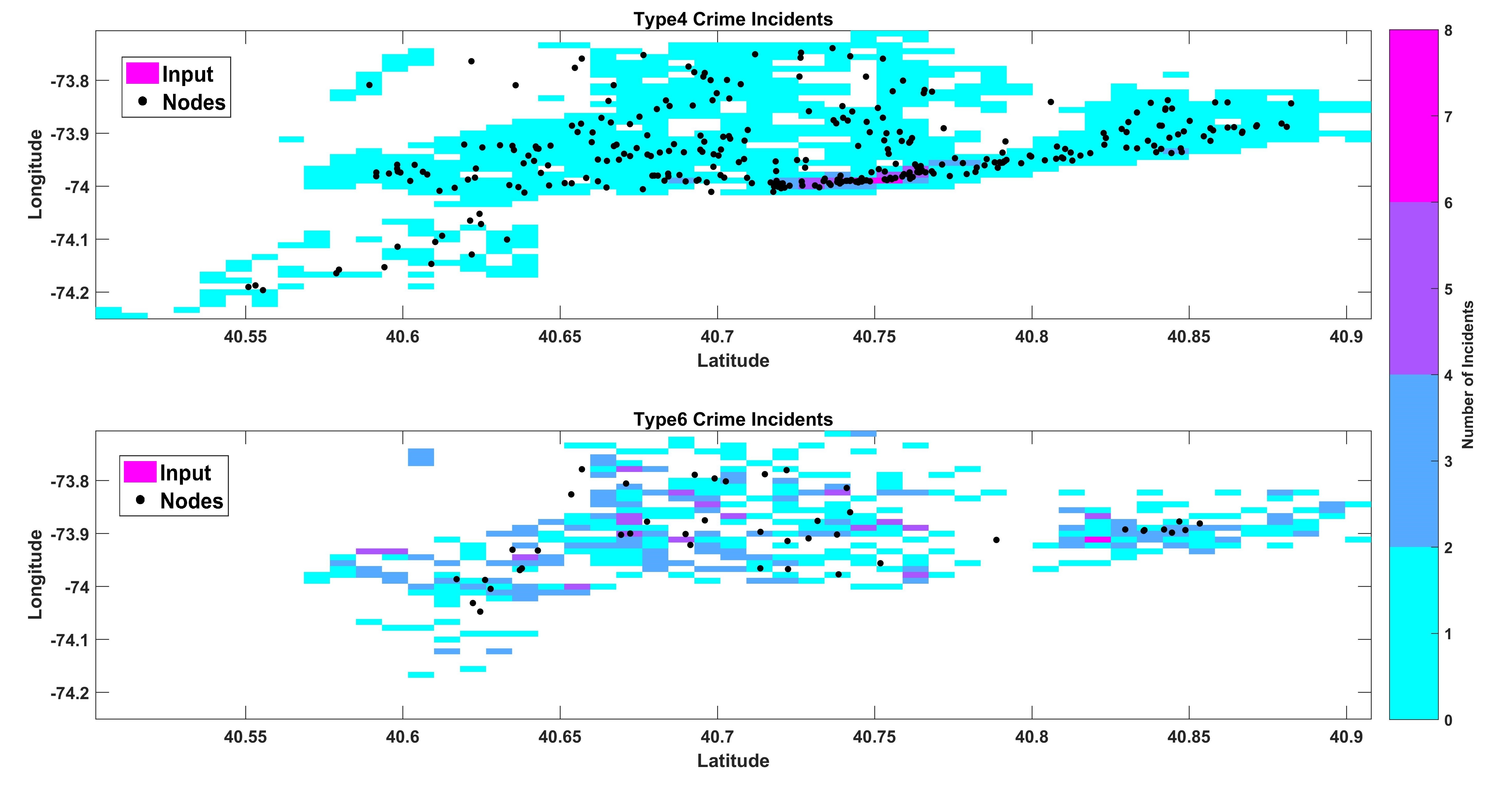}
		\caption{Distribution of Type 4 and 6 Felony}
	\end{center}
	\label{Fig:c4and6}
\end{figure}

\section{Discussion}
Generally, we could observe that CORs between nodes and input fluctuate more conspicuously, compared to that of hits and input. It can be explained by the fact that natural positions of layers along time axis are not always consistent to fixed equidistant division, sometimes it gave low CORs misleadingly, which is a limitation of this reliability measurement. Broader spans are set to reduce such side effect.
 
However, regardless of such drawback, 3D SOM performs well in basic clustering, for the CORs between hits and input remain at a high level, which are not troubled by such limitation. And the number of nodes matches with number of input in each layer. And according to Figure \ref{p2} and \ref{fig3}, nodes' distribution is consistent to input's density.

In multi-period analysis, results confirmed that 3D SOM could trace both daily and weekly pattern effectively. Beyond that, when projected into day-week plane, map discovers the correlation between two periods themselves. From Figure \ref{Fig:weekday}, it can be seen when crime incidents peaked changed as day passing by and nodes of map moved accordingly. It provides a new basis for multi-period analysis, that if split the nesting interval into matrix, much more information can be gathered together without loss of short terms details.

When it comes to categorical data, if we focus how the performance in reflecting each type, COR for each ID is strongly influenced by how many input belongs to that ID, for major types map fit well with input. However, because of the winner-take-all strategy, hardly any node is allocated to infrequent crimes with ID 1,7. Then I tried updating method based on the probability of each ID in weight matrix \textbf{W} , yet it fails to solve the problem and lead to decreased reliability. Detail on this method is supplied in \nameref{B2}. Linear regression analysis shows that it's almost inevitable due to contrast between sparsity of nodes and density of input.

While the results are promising in basic clustering, as the dimension increases, it's troublesome to judge whether a centroid locate at a namely right position. To be more concrete, when studying crime incidents, should the nodes move closer to input with same type of crime, or to data with less geographic distance? When adding ID　variables into clustering, CORs in temporal-spatial subspace decreased. Moreover, there are some unsolved theoretical problem\cite{rynkiewicz2006self} in self-organizing map itself\cite{unsolved}, prohibiting us to employ an universal standard to evaluate results.

Besides, this paper concentrates on patterns and relationship within the data, it's possible that temporal-spatial distributions are caused by factors not included, such as population, police force and so forth, leaving room for research to find potential causality.

\section*{Conclusion}
Three-dimensional self organizing map is competent and versatile in clustering and identifying behavioral patterns. Framework of grid reflects topographic feature of input, while codebook vectors store detailed information. Facilitated with an innovative methodology that expands timeline into matrix, 3D SOM unravels pattern and interrelation masked in multiple periods. Besides, a tailored 3D SOM clarifies complex relations in heterogeneous data, wisely avoiding calculation without analytical meaning.

In brief, what this paper seeks to address is not one or two specific clusterings of data contains time and geographic elements, but interpretation of interior relationships between variables based on temporal-spatial distribution. Self-organizing map is constructive for comprehension of high dimensional structures, providing an operable tool in functional utilization.

\section*{Appendix}
\subsection*{Appendix A}
\subsubsection*{Appendix A1}\label{A1}
\begin{flushleft}
	\textbf{Significance of Pre-Processing Input}
\end{flushleft}

Results shows that rescaling process can significantly reduce both the quantization and topographic error under the same training epochs. Their definitions are listed: \\
Rescaling the data to [0,1].
%Linear normalization
\begin{equation}
	x_{i}^* = \dfrac{x_{i}-x_{min}} {x_{max} - x_{min}}
\end{equation}
Standardization with z-score.
%Normalization
\begin{equation}
	x_{i}^* = \dfrac{x_{i} - \mu}{\sigma}
\end{equation} 
$\mu$ is the mean of $x$, and $\sigma$ represents standard deviation.\\
Following table contains QE and TE for data used in basic clustering trained with $10 \times 6 \times 6$ map after 100 epochs.
\begin{table}[h]
	\renewcommand{\arraystretch}{0.7}
	\caption{QE and TE of different types of normalization}
	\begin{center}
		\begin{tabular}{|c|c|c|c|}
			\hline
			Error Type & Raw Input & Rescaling & Z-score \\
			QE     &  0.0235   &  0.0251   & 0.0273  \\
			TE     &  0.4271   &  0.3595   & 0.3898  \\ \hline
		\end{tabular}
	\end{center}
\end{table}\\

Judging from the map plotted with different color representing each layer, normalized map meets less overlaps. Together with quantitative measurement above, we can conclude that normalization contributes to quality enhancement.
\begin{figure}[!h]
	%\nopagenumber
	%\renewcommand{\baselinestretch}{1.0}
	\hfill
	\begin{center}
		\includegraphics[width=5in]{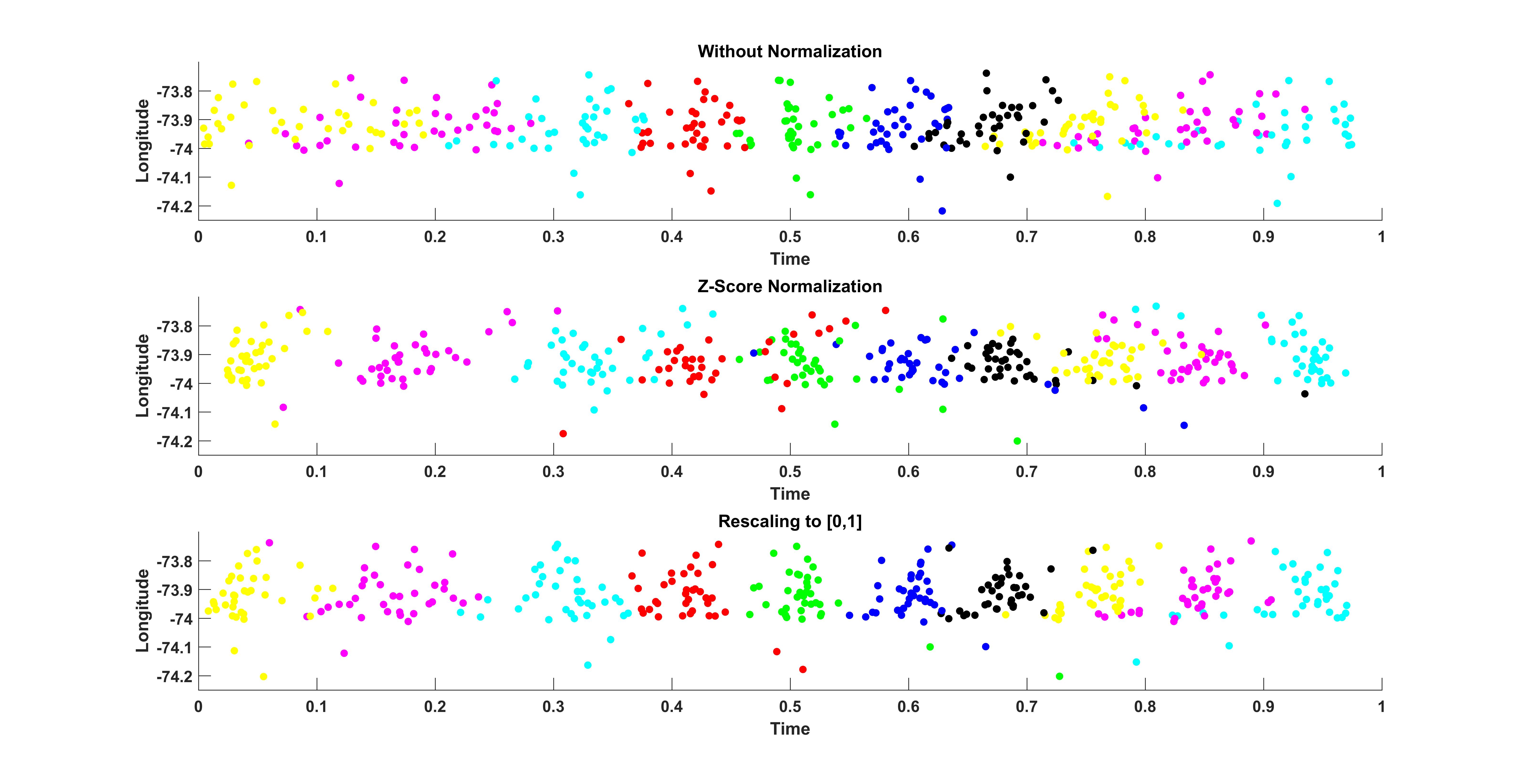}
		\caption{Influence of Normalization}
	\end{center}	
	\label{Fig:nor}
\end{figure}

\subsubsection*{Appendix A2} \label{A2}
\begin{flushleft}
	\textbf{Initialization of Map}
\end{flushleft}

Initialization process can be classified into random and linear initialization. While 2D SOM is known for being insensitive to different initialization, however for 3D SOM, if the codebook vector is initialized randomly, its topographic feature will remain chaotic with high error rate, which might be explained by metastability \cite{erwin1992self}.

Set map's size as $l \times m \times n$, where l is the number of layers along time axis and $m \times n$ is nodes in each layer. If the $max(l,m,n) \neq l$, map will rotate in the training, eventually the largest side of the 3D grid approximately parallels to time axis. The expected layered structure is lost, accompanied with high QE and TE. This results is consistent with the behavior of 2D SOM, where the two-dimensional grid extend itself in time-latitude or time-longitude plane, instead of latitude-longitude. It can be explained by Principle Component Analysis(PCA). It's found that time factor contributed the most to principle components. Therefore, in order to obtain less distortion and well organized layers, the size of grid should be chosen according to PCA, set the largest length to axis with greatest explaining ability.

\subsection*{Appendix B}
\subsubsection*{Appendix B1}\label{B1}
\begin{flushleft}
	\textbf{Algorithms for Categorical Variable}
\end{flushleft}

The updating of categorical variable is fully explained by following equation, inspired matrix calculations in SOM Toolbox\cite{vesanto1999self}.
\begin{equation*}
	\underset{m \times k}{W}  = \underset{m \times m}{M} \times (\underset{m \times n}{F} \times \underset{n \times k}{C})
\end{equation*}
\begin{center}
	The $m$,$n$,$k$ represent the number of nodes, inputs, and categories respectively.
\end{center}
\begin{itemize}
	\item Weight matrix \textbf{$W$}, $w_{i,j}$ indicate the weigh of category ${j}$ in $node_{i}$.
	\item Neighborhood matrix \textbf{$H$} measures the influence of $node_i$ to $node_j$. $H_{i,j}$ is calculated through Gaussian function $h(i,j)$,.
	\item Filter matrix \textbf{$F$}, indicating BMUs of input. If 1 appears in row $i$, column $j$, it means $input_i$'s BMU is $node_j$.
	\item Input ID matrix \textbf{$C$}, coding the categorical data into binaries, for row $i$, if column $j$ is 1, then ID variable in $input_i$ is $j$.
	\item The product of two matrices in parentheses indicates $node_i$ has $(F \times C)_{i,j}$ hits with category $j$.
\end{itemize}

\subsubsection*{Appendix B2}\label{B2}
\begin{flushleft}
	\textbf{Choose ID Variable in Probabilistic Manner}
\end{flushleft}

In probability based method for updating $m_{i}$'s ID variable, any ID values can be selected according to their probabilities.
\begin{eqnarray*}
	P(m_{i}=j) & = &P_{W_{i}}(j)\\
	P_{W_{i}}(j) & = & \dfrac{\sum\limits_{t}^n h_{c}n_j}{\sum\limits_{t}^n h_{c}}
\end{eqnarray*}
While it seems to be more considerate, results tells another story, that probability based method is affected by the negative impact of random numbers. Other research advices a combination of two method \cite{chen2005extension}, setting a threshold on minimum proportion for winner-take-all strategy, if it's not reached, selecting ID randomly.

\subsection*{Appendix C}
\subsubsection*{Appendix C1}\label{C1}
\begin{flushleft}
	\textbf{Additional Test on Multi-Period Analysis}
\end{flushleft}

Following is result of clustering 13717 vehicle collision reports\footnote{Jan.2015} with a $13 \times 8 \times 7$ map trained after 100 epochs. First, performance in one day is measured.
\begin{table}[!h]
	\renewcommand{\arraystretch}{0.7}	
	\caption{CORs between Input and Map in Days}
	\begin{center}
		\begin{tabular}{|c|c|c|c|c|c|c|c|c|}
			\hline
			Input \& Nodes &0.73& 0.86& 0.93& 0.92& 0.97& 0.94& 0.87& 0.23\\
			Inputs \& Hits &0.94& 0.92& 0.99& 0.99& 0.99& 0.98& 0.98& 0.93\\ \hline
		\end{tabular}
	\end{center}
\end{table}
\begin{table}[!h]
	\renewcommand{\arraystretch}{0.7}	
	\caption{Daily Sum of Input and Map}
	\begin{center}
		\begin{tabular}{|c|c|c|c|c|c|c|c|c|}
			\hline
			Input & 731 & 510 & 2060 & 2481 & 2812 & 2650 & 1776 & 693 \\		
			Hits  & 738 & 480 & 2193 & 2463 & 2754 & 2670 & 1717 & 698 \\ 
			Nodes & 45  & 35  & 109  & 127  & 143  & 137  &  92  & 40  \\\hline
		\end{tabular}
	\end{center}
\end{table}\\

Then, using same measurement on performance in one week.
\begin{table}[!h]
	\renewcommand{\arraystretch}{0.7}	
	\caption{Coefficients between Input and Map in Week}
	\begin{center}
		\begin{tabular}{|c|c|c|c|c|c|c|c|}
			\hline
			Day in Week & Mon. & Tue. & Wed. & Thur. & Fri. & Sat. & Sun. \\ \hline
			Nodes    & 0.72 & 0.91 & 0.87 & 0.86  & 0.86 & 0.69 & 0.78 \\
			Hits     & 0.98 & 0.98 & 0.98 & 0.98  & 0.98 & 0.97 & 0.98 \\ \hline
		\end{tabular}
	\end{center}
\end{table}
\begin{table}[!h]
	\renewcommand{\arraystretch}{0.7}	
	\caption{Weekly Sum of Input and Map}
	\begin{center}
		\begin{tabular}{|c|c|c|c|c|c|c|c|}
			\hline
			Day in Week & Mon. & Tue. & Wed. & Thur. & Fri. & Sat. & Sun. \\ \hline
			Input Data  & 1739 & 1743 & 1713 & 2178  & 2555 & 2039 & 1746 \\
			SOM Hits   & 1731 & 1749 & 1726 & 2176  & 2546 & 1998 & 1787 \\
			SOM Nodes  &  96  & 105  & 105  &  106  & 113  & 114  &  89  \\ \hline
		\end{tabular}
	\end{center}
\end{table}\\
\subsubsection*{Appendix C2}\label{C2}
\begin{flushleft}
	\textbf{ Details in Performance on Heterogeneous Input}
\end{flushleft}

Following table is classification of crimes with IDs.
\begin{table}[!h]
	\label{Felonies}
	\renewcommand{\arraystretch}{0.7}
	\caption{ID Numbers of Felonies}
	\begin{center}
		\begin{tabular}{|c|c|}
			\hline
			ID Number &          Offense Type          \\ \hline
			1     &              Rape              \\
			2     &            Burglary            \\
			3     &         Felony Assault         \\
			4     &         Grand Larceny          \\
			5     &            Robbery             \\
			6     & Grand Larceny of Motor Vehicle \\
			7     &  Murder Non-Negl.Manslaughter  \\ \hline
		\end{tabular}
	\end{center}
\end{table}\\

Table \ref{tab:15} and \ref{tab:16} supplied clustering of temporal-spatial distribution of all crimes.\\
\begin{table}[!h]
	\renewcommand{\arraystretch}{0.7}	
	\caption{Coefficients between Input and Map in Day}\label{tab:15}
	\begin{center}
		\begin{tabular}{|c|c|c|c|c|c|c|c|c|}
			\hline
			Input \& Nodes & -0.14 & 0.87 & 0.73 & 0.73 & 0.96 & 0.93 & 0.86 & 0.47 \\
			Inputs \& Hits & 0.98  & 0.95 & 0.90 & 0.97 & 0.98 & 0.99 & 0.97 & 0.94 \\ \hline
		\end{tabular}
	\end{center} 
\end{table}

\begin{table}[!h]
	\renewcommand{\arraystretch}{0.7}	
	\caption{Daily Sum of Input and Map}\label{tab:16}
	\begin{center}
		\begin{tabular}{|c|c|c|c|c|c|c|c|c|c|}
			\hline
			Input & 976 & 540 & 729 & 1041 & 1239 & 1380 & 1244 & 667 & $R^{2}$ \\
			Nodes & 64  & 54  & 89  & 109  & 124  & 136  & 110  & 42  &  0.77   \\
			Hits  & 917 & 552 & 851 & 1064 & 1169 & 1365 & 1257 & 641 &  0.96   \\ \hline
		\end{tabular}
	\end{center}
\end{table}   
\qquad \\

\section*{References}

\bibliography{mybibfile}

\end{document}